# Multi-valued Color Representation Based on Frank t-norm Properties

Vasile Patrascu

Department of Informatics Technology, Tarom Company
Bucurestilor Road, 224F, Bucharest-Otopeni, Romania
e-mail: patrascu.v@gmail.com

### Abstract

In this paper two knowledge representation models are proposed, FP4 and FP6. Both combine ideas from fuzzy sets and four-valued and hexa-valued logics. Both represent imprecise properties whose accomplished degree is unknown or contradictory for some objects. A possible application in the color analysis and color image processing is discussed.

## 1    Introduction

Fuzzy sets are a specially well-suited tool to represent imprecise concepts with ill-defined boundaries. When a property $P$ is imprecise, its negation $\neg P$ is considered to be imprecise. The fuzzy set theory assumes both $P$ and $\neg P$ are related, namely: $\neg P(x) = 1 - P(x)$. However, this is not always true in real life. Hence, sometimes $P$ and $\neg P$ are represented independently. On the other hand, fuzzy sets do not allow to take into account the presence of objects whose membership degree $P$ is unknown. Three valued logics can solve the problem allowing three truth values: true, false and unknown. However, this is not always sufficient and we can find contradictions when a certain value $x$ verifies $P$ and $\neg P$ at the same time. Four value logics can solve the problem because it uses four truth values: true, false, unknown and contradictory. There are two special situations when the property $P$ is described by two other properties $P_1$ and $P_2$, namely $P(x) = P_1(x) \cap P_2(x)$; alternatively the negation is described by the negation of two other properties $\neg P_1$ and $\neg P_2$, namely $\neg P(x) = \neg P_1(x) \cup \neg P_2(x)$. In this situation, we can detail the knowledge representation using a hexa-valued logic based on six truth values: certain-true, uncertain-true, certain-false, uncertain-false, unknown and contradictory. In conclusion, the paper proposes two knowledge representation models, where fuzzy sets and four-valued and hexa-valued logics are combined to represent imprecise properties.

## 2    The Generalized Fuzzy Sets

Let $X$ be a crisp set. In the framework of the Zadeh theory [12], a fuzzy set $A$ is defined by the membership function $\mu_A : X \to [0,1]$. The non-membership function $\nu_A : X \to [0,1]$ is obtained by negation and thus both functions define a partition of unity, namely:

$$\mu_A + \nu_A = 1 \qquad (2.1)$$

Atanassov has extended the fuzzy sets to the intuitionistic fuzzy sets [1]. Atanassov has relaxed the condition (2.1) to the following inequality:

$$\mu_A + \nu_A \leq 1 \qquad (2.2)$$

He has used the third function, the index of uncertainty $\pi_A$ that verifies the equality:

$$\mu_A + \nu_A + \pi_A = 1 \qquad (2.3)$$

In this way, the set $A$ is characterized by a three-valued partition of unity. Belnap has defined a four-valued logic based on true, false, uncertainty and contradiction [2]. Thus, using Belnap's logic, we can define a new type of set

1215

by constructing a four-valued partition. Moreover, we broaden this four-valued partition to a hexa-valued one, by replacing the membership and non-membership functions with four independent functions: strong membership, weak membership, strong non-membership and weak non-membership [9], [10], [11].

In this paper we will consider as bipolar fuzzy set (BFS), a set $A$ defined by the functions $\mu_A : X \to [0,1]$, $\nu_A : X \to [0,1]$ and these two functions are totally independently.

## 3 Transformations from BFS to Four-valued and Hexa-valued Fuzzy Partitions

Let there be the Frank t-norm [4] defined for $s \in (0,\infty)$ by:

$$t_s(x,y) = \log_s\left(1 + \frac{(s^x - 1)\cdot(s^y - 1)}{s - 1}\right) \quad (3.1)$$

Let there be a Frank t-norm denoted by „∘". This t-norm verifies the Frank equation [4]:

$$x \circ y - \bar{x} \circ \bar{y} = x + y - 1 \quad (3.2)$$

where $\bar{x}$ is the negation of $x$, namely:

$$\bar{x} = 1 - x$$

An equivalent form of Frank equation one can obtain by replacing $y$ with $(1-y)$, namely:

$$x \circ \bar{y} - y \circ \bar{x} = x - y \quad (3.3)$$

Also, one defines its dual or its t-conorm „⊕" by:

$$x \oplus y = 1 - \bar{x} \circ \bar{y}$$

and thus, the formula (3.2) has the equivalent form:

$$x \oplus y + x \circ y = x + y$$

Let there be two t-norms „∘" and „•". We say that these two t-norms are *conjugated* if for $\alpha + \beta = 1$ there exists the equality:

$$x = x \bullet \alpha + x \circ \beta \quad (3.4)$$

Immediately, one results:

$$x \bullet y = x - x \circ \bar{y} \quad (3.5)$$

$$x \circ y = x - x \bullet \bar{y} \quad (3.6)$$

From (3.1) and (3.5) one obtains for the conjugate:

$$x \bullet y = \log_{\frac{1}{s}}\left(1 + \frac{\left(\frac{1}{s^x} - 1\right)\cdot\left(\frac{1}{s^y} - 1\right)}{\frac{1}{s} - 1}\right)$$

Thus, two Frank t-norms are *conjugated* if one is computed with parameter $s$ and the other is computed with parameter $\frac{1}{s}$. Thus, it results that the logics Godel and Lukasiewicz [6], [7[, [8] are conjugated and the Product logic is identically with its conjugate.

For $s = 0$ it results:

$$\begin{cases} x \oplus y = Max(x,y) \\ x \circ y = Min(x,y) \\ x \bullet y = Max(0, x+y-1) \end{cases} \quad (3.7)$$

From (3.5) one obtains

$$x = x \bullet \bar{y} + x \circ y \quad (3.8)$$

and replacing $y$ by $\bar{y}$ it results:

$$\bar{x} = \bar{x} \bullet y + \bar{x} \circ \bar{y} \quad (3.9)$$

We denote:

$$\begin{cases} \tau = x \bullet \bar{y} \\ \varphi = \bar{x} \bullet y \\ \pi = \bar{x} \circ \bar{y} \\ \kappa = x \circ y \end{cases} \quad (3.10))$$

Adding (3.8) with (3.9) one obtains a partition of unity, namely:

$$1 = \tau + \varphi + \kappa + \pi \quad (3.11)$$

The formula (3.11) will base the four-valued fuzzy partition. Replacing $y$ by $y \oplus z$ in (3.8), (3.9) it results:

$$\begin{cases} x = x \bullet \overline{y \oplus z} + x \circ (y \oplus z) \\ \bar{x} = \bar{x} \bullet y \oplus z + \bar{x} \circ \overline{y \oplus z} \end{cases}$$

or

$$\begin{cases} x = x \bullet \bar{y} \circ \bar{z} + (x \circ (y \oplus z) - x \circ y \circ z) + x \circ y \circ z \\ \bar{x} = \bar{x} \bullet (y \circ z) + (\bar{x} \bullet y \oplus z - \bar{x} \bullet (y \circ z)) + \bar{x} \circ \bar{y} \circ \bar{z} \end{cases}$$

$$(3.12)$$

Now, we will denote:

$$\begin{cases} \tau = x \bullet \overline{y} \circ \overline{z} \\ \lambda = x \circ (y \oplus z) - x \circ y \circ z \\ \kappa = x \circ y \circ z \\ \pi = \overline{x} \circ \overline{y} \circ \overline{z} \\ \omega = \overline{x} \bullet (y \oplus z) - \overline{x} \bullet (y \circ z) \\ \varphi = \overline{x} \bullet (y \circ z) \end{cases} \quad (3.13)$$

From (3.12) and (3.13) it results a new partition of unity, namely:

$$1 = \tau + \varphi + \kappa + \pi + \lambda + \omega \quad (3.14)$$

Now, replacing $y$ by $y \circ z$ from (3.8), (3.9) it results:

$$\begin{cases} x = x \bullet \overline{y \circ z} + x \circ y \circ z \\ \overline{x} = \overline{x} \bullet (y \circ z) + \overline{x} \circ \overline{y \circ z} \end{cases}$$

or

$$\begin{cases} x = x \bullet (\overline{y} \circ \overline{z}) + (x \bullet (\overline{y} \oplus \overline{z}) - x \bullet (\overline{y} \circ \overline{z})) + x \circ y \circ z \\ \overline{x} = \overline{x} \bullet (y \circ z) + (\overline{x} \circ (\overline{y} \oplus \overline{z}) - \overline{x} \circ \overline{y} \circ \overline{z}) + \overline{x} \circ \overline{y} \circ \overline{z} \end{cases} \quad (3.15)$$

Let us denote:

$$\begin{cases} \tau = (y \circ z) \bullet \overline{x} \\ \lambda = \overline{x} \circ (\overline{y} \oplus \overline{z}) - \overline{x} \circ \overline{y} \circ \overline{z} \\ \kappa = x \circ y \circ z \\ \pi = \overline{x} \circ \overline{y} \circ \overline{z} \\ \omega = x \bullet (\overline{y} \oplus \overline{z}) - x \bullet (\overline{y} \circ \overline{z}) \\ \varphi = (\overline{y} \circ \overline{z}) \bullet x \end{cases} \quad (3.16)$$

We have again obtained a hexa-valued partition of unity because the equality (3.14) is true. The formulae (3.13) and (3.16) will base the hexa-valued fuzzy partition.

We consider the set $A \in BFS$ having the membership function $\mu_A$ and non-membership function $\nu_A$. Using (3.10) we define $\tau_A, \varphi_A, \pi_A, \kappa_A$ the indexes of truth, falsity, contradiction and uncertainty.

$$\begin{cases} \tau_A = \mu_A \bullet (1 - \nu_A) \\ \varphi_A = (1 - \mu_A) \bullet \nu_A \\ \pi_A = (1 - \mu_A) \circ (1 - \nu_A) \\ \kappa_A = \mu_A \circ \nu_A \end{cases} \quad (3.17)$$

We have transformed a binary representation of knowledge in a four-valued one. From (3.2) and (3.3) the following two equalities result:

$$\begin{cases} \mu_A - \nu_A = \tau_A - \varphi_A \\ \mu_A + \nu_A = 1 + \kappa_A - \pi_A \end{cases} \quad (3.18)$$

Immediately, one obtains the inverse transform:

$$\begin{cases} \mu_A = \dfrac{1 + \kappa_A - \pi_A + \tau_A - \varphi_A}{2} \\ \nu_A = \dfrac{1 + \kappa_A - \pi_A - \tau_A + \varphi_A}{2} \end{cases} \quad (3.19)$$

If the non-membership function has the particular form:

$$\nu_A = \nu_A^1 \oplus \nu_A^2$$

then, using (3.13), it results:

$$\begin{cases} \tau_A = \mu_A \bullet \overline{\nu}_A^1 \circ \overline{\nu}_A^2 \\ \lambda_A = \mu_A \circ (\nu_A^1 \oplus \nu_A^2) - \mu_A \circ \nu_A^1 \circ \nu_A^2 \\ \kappa_A = \mu_A \circ \nu_A^1 \circ \nu_A^2 \\ \pi_A = \overline{\mu}_A \circ \overline{\nu}_A^1 \circ \overline{\nu}_A^2 \\ \omega_A = \overline{\mu}_A \bullet (\nu_A^1 \oplus \nu_A^2) - \overline{\mu}_A \bullet (\nu_A^1 \circ \nu_A^2) \\ \varphi_A = \overline{\mu}_A \bullet (\nu_A^1 \circ \nu_A^2) \end{cases} \quad (3.20)$$

We have obtained a hexa-valued knowledge representation where $\tau, \lambda, \kappa, \pi, \omega, \varphi$ represent in order: the strong membership, the weak membership, the contradiction, the uncertainty, the weak non-membership and the strong non-membership.

If the membership function has the following form:

$$\mu_A = \mu_A^1 \circ \mu_A^2$$

then, using (3.16), it results:

$$\begin{cases} \tau_A = (\mu_A^1 \circ \mu_A^2) \bullet \overline{\nu}_A \\ \lambda_A = (\overline{\mu}_A^1 \oplus \overline{\mu}_A^2) \circ \overline{\nu}_A - \overline{\mu}_A^1 \circ \overline{\mu}_A^2 \circ \overline{\nu}_A \\ \kappa_A = \mu_A^1 \circ \mu_A^2 \circ \nu_A \\ \pi_A = \overline{\mu}_A^1 \circ \overline{\mu}_A^2 \circ \overline{\nu}_A \\ \omega_A = (\overline{\mu}_A^1 \oplus \overline{\mu}_A^2) \bullet \nu_A - (\overline{\mu}_A^1 \circ \overline{\mu}_A^2) \bullet \nu_A \\ \varphi_A = (\overline{\mu}_A^1 \circ \overline{\mu}_A^2) \bullet \nu_A \end{cases} \quad (3.21)$$

Also, the formulae (3.21) define a hexa-valued knowledge representation.

## 4 Four-valued and Hexa-valued Fuzzy Partition for Color Space

In this section we will construct four-valued and hexa-valued fuzzy partitions for some color properties description. We consider as color space the *RGB* system (*red*, *green* and *blue*) [3], [5]. We will suppose that $R, G, B \in [0,1]$.

Firstly, we define the bipolar fuzzy sets for the following color properties:

redness:
$$\begin{cases} \mu_R = R \\ \nu_R = B \oplus G \end{cases} \quad (4.1)$$

greenness:
$$\begin{cases} \mu_G = G \\ \nu_G = R \oplus B \end{cases} \quad (4.2)$$

blueness:
$$\begin{cases} \mu_B = B \\ \nu_B = R \oplus G \end{cases} \quad (4.3)$$

yellowness:
$$\begin{cases} \mu_Y = R \circ G \\ \nu_Y = B \end{cases} \quad (4.4)$$

magentaness:
$$\begin{cases} \mu_M = R \circ B \\ \nu_M = G \end{cases} \quad (4.5)$$

cyanness:
$$\begin{cases} \mu_C = B \circ G \\ \nu_C = R \end{cases} \quad (4.6)$$

whiteness:
$$\begin{cases} \mu_W = R \circ G \circ B \\ \nu_W = 1 - R \circ G \circ B \end{cases} \quad (4.7)$$

blackness:
$$\begin{cases} \mu_K = 1 - R \oplus G \oplus B \\ \nu_K = R \oplus G \oplus B \end{cases} \quad (4.8)$$

brightness:
$$\begin{cases} \mu_H = R \circ G \circ B \\ \nu_H = 1 - R \oplus G \oplus B \end{cases} \quad (4.9)$$

darkness:
$$\begin{cases} \mu_L = 1 - R \oplus G \oplus B \\ \nu_L = R \circ G \circ B \end{cases} \quad (4.10)$$

Now, for redness one obtains the following four-valued and hexa-valued representations using (3.17) and (3.20):

$$\begin{cases} \tau_R = R \bullet (\overline{B} \circ \overline{G}) \\ \varphi_R = \overline{R} \bullet (B \oplus G) \\ \pi_R = \overline{R} \circ \overline{B} \circ \overline{G} \\ \kappa_R = R \circ (B \oplus G) \end{cases} \quad (4.11)$$

$$\begin{cases} \tau_R = R \bullet \overline{G} \circ \overline{B} \\ \lambda_R = R \circ (G \oplus B) - R \circ G \circ B \\ \kappa_R = R \circ G \circ B \\ \pi_R = \overline{R} \circ \overline{G} \circ \overline{B} \\ \omega_R = \overline{R} \bullet (G \oplus B) - \overline{R} \bullet (G \circ B) \\ \varphi_R = \overline{R} \bullet (G \circ B) \end{cases} \quad (4.12)$$

For blueness and greenness one obtains similar formulae.

For yellowness we obtain the following four-valued fuzzy partition:

$$\begin{cases} \tau_Y = (R \circ G) \bullet \overline{B} \\ \varphi_Y = (\overline{R} \oplus \overline{G}) \bullet B \\ \pi_Y = (\overline{R} \oplus \overline{G}) \circ \overline{B} \\ \kappa_Y = R \circ G \circ B \end{cases} \quad (4.13)$$

A hexa-valued fuzzy partition results from formulae (3.21).

$$\begin{cases} \tau_Y = (R \circ G) \bullet \overline{B} \\ \lambda_Y = (\overline{R} \oplus \overline{G}) \circ \overline{B} - \overline{R} \circ \overline{G} \circ \overline{B} \\ \kappa_Y = R \circ G \circ B \\ \pi_Y = \overline{R} \circ \overline{G} \circ \overline{B} \\ \omega_Y = (\overline{R} \oplus \overline{G}) \bullet B - (\overline{R} \circ \overline{G}) \bullet B \\ \varphi_Y = (\overline{R} \circ \overline{G}) \bullet B \end{cases} \quad (4.14)$$

For magentaness and cyanness one obtains similar formulae.

Using (3.17) we obtain the four-valued fuzzy representation for whiteness:

$$\begin{cases} \tau_W = (R \circ G \circ B) \bullet (R \circ G \circ B) \\ \varphi_W = (\overline{R} \oplus \overline{G} \oplus \overline{B}) \bullet (\overline{R} \oplus \overline{G} \oplus \overline{B}) \\ \pi_W = (\overline{R} \oplus \overline{G} \oplus \overline{B}) \circ R \circ G \circ B \\ \kappa_W = R \circ G \circ B \circ (\overline{R} \oplus \overline{G} \oplus \overline{B}) \end{cases} \quad (4.15)$$

For blackness we obtain the following four-valued representation:

$$\begin{cases} \tau_K = (\overline{R} \circ \overline{G} \circ \overline{B}) \bullet (\overline{R} \circ \overline{G} \circ \overline{B}) \\ \varphi_K = (R \oplus G \oplus B) \bullet (R \oplus G \oplus B) \\ \pi_K = (R \oplus G \oplus B) \circ (\overline{R} \circ \overline{G} \circ \overline{B}) \\ \kappa_K = (\overline{R} \circ \overline{G} \circ \overline{B}) \circ (R \oplus G \oplus B) \end{cases} \quad (4.16)$$

The blackness is the complement of the whiteness. Analyzing the formulae (4.12) we remark that the parameter $\tau_R$ shows that the analyzed color is reddish, $\varphi_R$ shows that the color is not reddish and is close to cyan, $\kappa_R$ shows that the color is bright, $\pi_R$ shows that the color is dark, $\lambda_R$ shows that the color is close to yellow or to magenta and $\omega_R$ shows that the color is close to blue or green. Analyzing formulae (4.15) we observe that $\tau_W$ shows that the color is close to white while $\varphi_W$ shows that the color is close to black. The parameters $\kappa_W$ and $\pi_W$ show that the color is unsaturated and is close to the middle gray. Taking into account only the truth functions we can define the following two 8-parameter descriptors:

$$V_{wk} = (\tau_R, \tau_Y, \tau_G, \tau_C, \tau_B, \tau_M, \tau_W, \tau_K) \quad (4.17)$$

$$V_{hl} = (\tau_R, \tau_Y, \tau_G, \tau_C, \tau_B, \tau_M, \tau_H, \tau_L) \quad (4.18)$$

The coordinate system (4.17) is less sensible to the color luminosity variations and it can be used by the segmentation procedure.

Also, we must underline that the vectors (4.17) and (4.18) do not define partitions of unity but the sum of their components is less than one. One defines for each of them the index of color indeterminacy by formulae:

$$i_{wk} = 1 - \tau_R - \tau_Y - \tau_G - \tau_C - \tau_B - \\ - \tau_M - \tau_W - \tau_K \quad (4.19)$$

$$i_{hl} = 1 - \tau_R - \tau_Y - \tau_G - \tau_C - \tau_B - \\ - \tau_M - \tau_H - \tau_L \quad (4.20)$$

Moreover, if $R,G,B$ are the coordinates of RGB cube center then all the components of the vectors defined by (4.17) and (4.18) are zero.

In the case of the pair of logics Lukasiewicz-Godel (3.7) one obtains the following particular forms for the parameters considered in (4.17), (4.18), (4.19) and (4.20):

$$\begin{cases} \tau_R = (R - \max(B,G))_+ \\ \tau_Y = (\min(R,G) - B)_+ \\ \tau_G = (G - \max(B,R))_+ \\ \tau_C = (\min(G,B) - R)_+ \\ \tau_B = (B - \max(R,G))_+ \\ \tau_M = (\min(R,B) - G)_+ \end{cases} \quad (4.21)$$

$$\begin{cases} \tau_W = (2 \cdot \min(R,G,B) - 1)_+ \\ \tau_K = (1 - 2 \cdot \max(R,G,B))_+ \end{cases} \quad (4.22)$$

$$\begin{cases} \tau_H = (\min(R,G,B) + \max(R,G,B) - 1)_+ \\ \tau_L = (1 - \min(R,G,B) - \max(R,G,B))_+ \end{cases} \quad (4.23)$$

$$i_{wk} = 1 - |\min(R,G,B) - 0.5| - \\ - |\max(R,G,B) - 0.5| \quad (4.24)$$

$$i_{hl} = 1 - \max(R,G,B) + \min(R,G,B) - \\ - |\min(R,G,B) + \max(R,G,B) - 1| \quad (4.25)$$

where $x_+$ is the positive part of $x$, namely:

$$x_+ = \frac{x + |x|}{2} \quad (4.26)$$

## 5 Experimental Results

In figures (fig. 1-6), (fig. 7-12), (fig. 13-18) and (fig. 19-24) one presents the RGB values, the hexa-valued partitions and the eight-parameter vectors defined by formulae (4.17), (4.18) for the following colors: (1.0,0.5,0.8), (0.5,0.0,0.3), (0.1,0.4,0.9) and (0.9,0.6,0.1). The first two have the same hue and saturation but one is bright and the other is dark. The last two colors are complementary.

In figures (fig. 25-30) one presents the histogram generated by the RGB values, the fuzzy cardinalities of hexa-valued partitions and vector component defined by (4.17), (4.18) calculated for the image "Girl" (fig. 25).

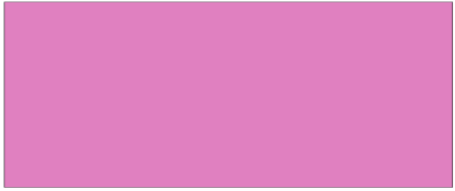

Figure 1: The color (1,0.5,0.8).

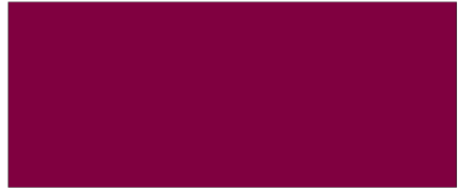

Figure 7: The color (0.5,0,0.3).

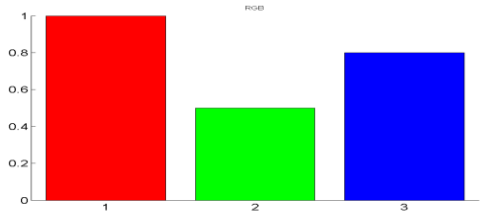

Figure 2: The RGB structure.

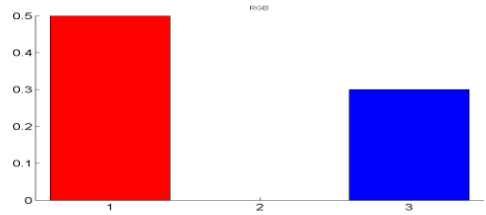

Figure 8: The RGB structure.

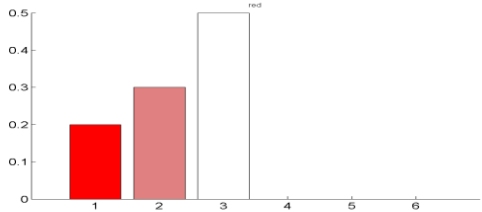

Figure 3: The redness structure.

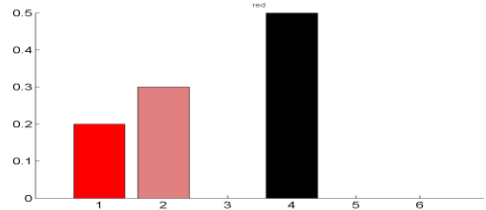

Figure 9: The redness structure.

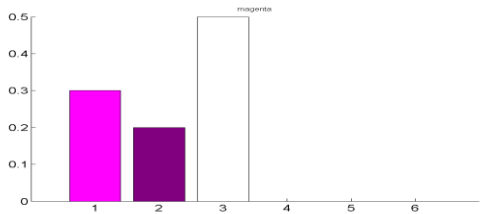

Figure 4: The magentaness structure.

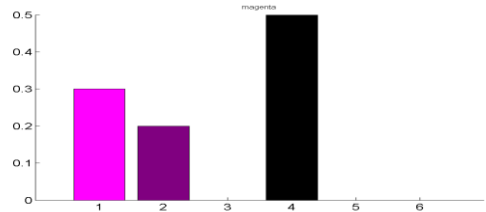

Figure 10: The magentaness structure.

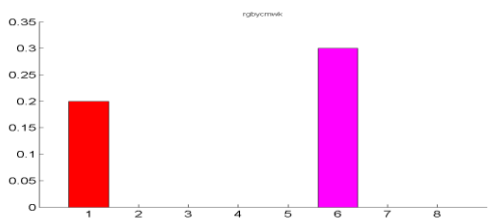

Figure 5: The rgbycmwk structure.

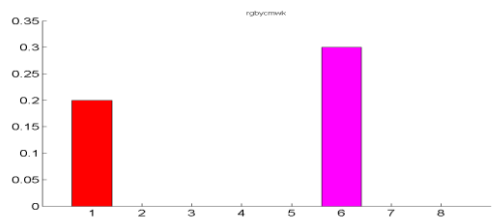

Figure 11: The rgbycmwk structure.

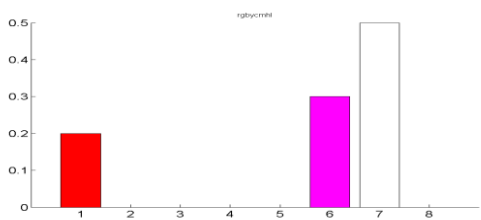

Figure 6: The rgbycmhl structure.

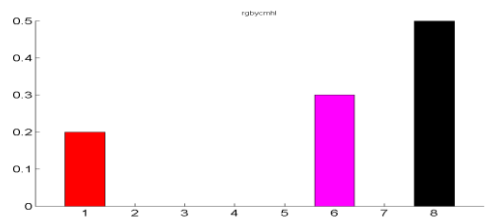

Figure 12: The rgbycmhl structure.

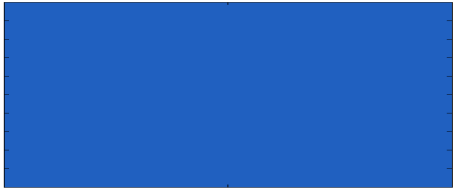

Figure 13: The color (0.1,0.4,0.9).

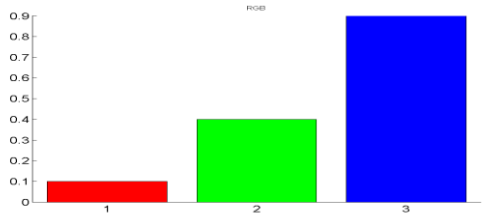

Figure 14: The RGB structure.

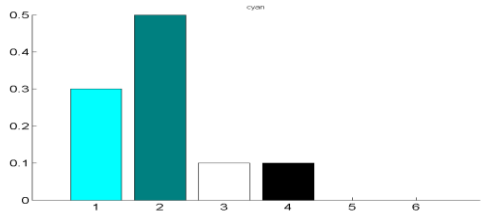

Figure 15: The cyanness structure.

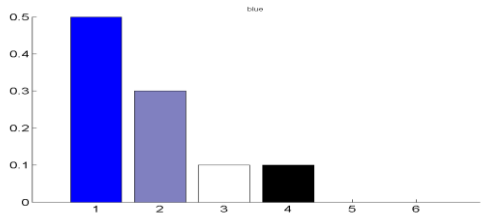

Figure 16: The blueness structure.

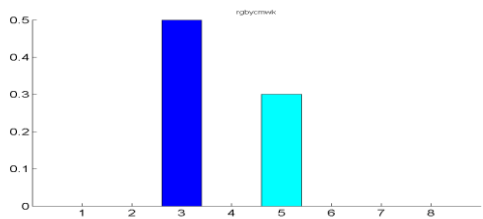

Figure 17: The rgbycmwk structure.

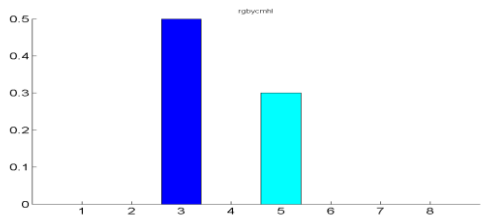

Figure 18: The rgbycmhl structure.

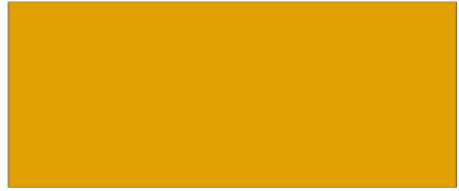

Figure 19: The color (0.9,0.6,0.1).

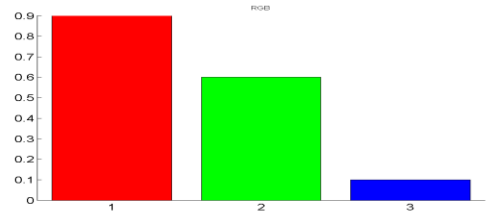

Figure 20: The RGB structure.

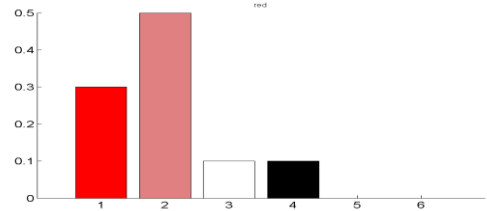

Figure 21: The redness structure.

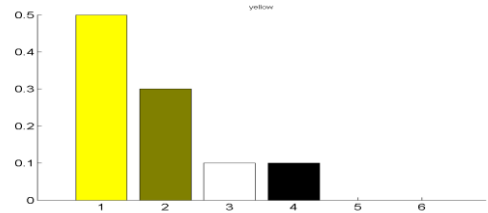

Figure 22: The yellowness structure.

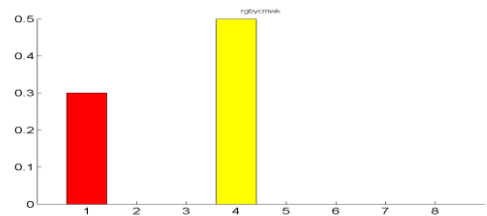

Figure 23: The rgbycmwk structure.

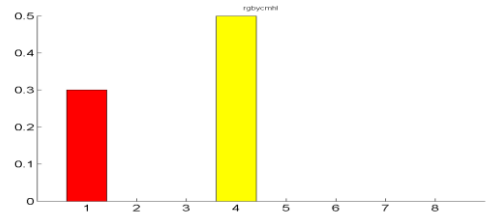

Figure 24: The rgbycmhl structure.

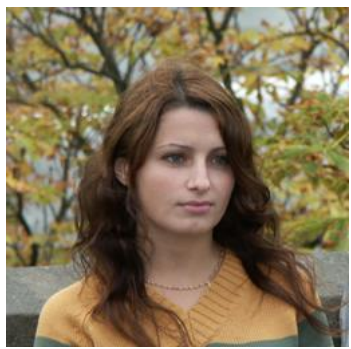

Figure 25: The image "Girl".

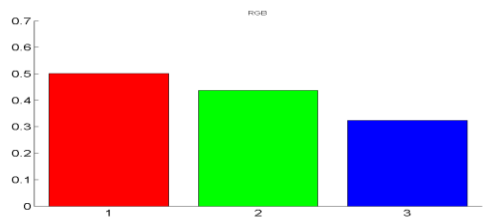

Figure 26: The RGB histogram.

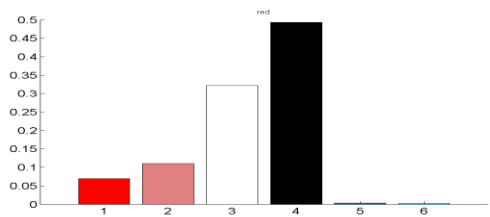

Figure 27: The redness histogram.

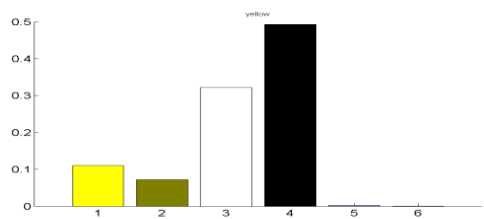

Figure 28: The yellowness histogram.

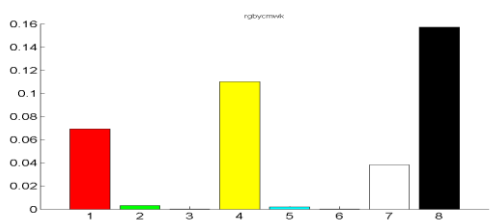

Figure 29: The rgbycmwk histogram.

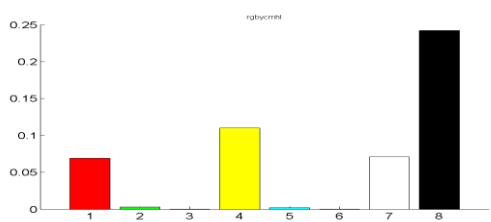

Figure 30: The rgbycmhl histogram.

## 6 Conclusions

This paper presented some new methods regarding the multi-valued representation of knowledge and their applications in color analysis domain. The methods are based on some properties of the Frank t-norms. Another important thing is the definition of bipolar fuzzy sets in RGB color space.

## References

[1] K. Atanassov (1986). Intuitionistic fuzzy sets, *Fuzzy Sets and Systems 20*, pp. 87-96, 1986.

[2] N. Belnap (1977). A Useful Four-valued Logic, *Modern Uses of Multiple-valued Logics*, Dordrecht-Boston, pp. 8-37, 1977.

[3] K. R. Castleman (1996). *Digital Image Processing*. Prentice Hall, Englewood Cliffs Nj, 1996.

[4] M. J. Frank (1979). On the simultaneous associativity of f(x,y) and x+y-f(x,y). *Aeq. Math.*, 19:194-226, 1979.

[5] A. K. Jain (1989). *Fundamentals of Digital Image Processing.* Prentice Hall, Englewood Cliffs Nj, 1989.

[6] Gr. C. Moisil (1965). *Old and New essays on non-classical logics* (Romanian), Ed. Stiintifica, Bucharest, 1965.

[7] Gr. C. Moisil (1972). *Essai sur les logiques non-chrysippiennes*, Ed. Academiei, Bucharest, 1972.

[8] Gr. C. Moisil (1975). *Lectures on the logic of fuzzy reasoning,* Ed. Stiintifica, Bucharest, 1975.

[9] V. Patrascu (2007). Penta-Valued Fuzzy Set, *The IEEE International Conference on Fuzzy Systems,* London, pp. 137-140, U.K, July 24-26, 2007.

[10] V. Patrascu (2007). Rough Sets on Four-Valued Fuzzy Approximation Space, *The IEEE International Conference on Fuzzy Systems*, London, pp. 217-221, U.K, July 24-26, 2007.

[11] V. Patrascu (2006). Fuzzy set based on four valued fuzzy logic, *The International Conference on Computers, Communications & Control*, (ICCCC 2006), pp. 360-365, Baile Felix - Oradea, Romania, 2006.

[12] L. A. Zadeh (1965). Fuzzy sets, *Information and Control*, 8, pp. 338-353, 1965.